# Super Resolution for Root Imaging[1]

Jose F. Ruiz-Munoz,[2] Jyothier K. Nimmagadda,[3] Tyler G. Dowd,[4] James E. Baciak[3] and Alina Zare,[2, 5]

[2] University of Florida, Department of Electrical and Computer Engineering, Gainesville, FL, USA

[3] University of Florida, Department of Material Sciences and Engineering, Gainesville, FL, USA

[4] Donald Danforth Plant Science Center, Saint Louis, Missouri, USA

Email addresses: JFR: jruizmunoz@ufl.edu

JKN: jyothir@ufl.edu

TGD: tdowd@danforthcenter.org

JEB: jebaciak@mse.ufl.edu

AZ: azare@ece.ufl.edu

Number of words: 3844

## ABSTRACT

- *Premise of the study*: High-resolution cameras have become very helpful for plant phenotyping by providing a mechanism for tasks such as target versus background discrimination, and the measurement and analysis of fine-above-ground plant attributes. However, the acquisition of high-resolution (HR) imagery of plant roots is more challenging than above-ground data collection. Thus, an effective super-resolution (SR) algorithm is desired for overcoming resolution limitations of sensors, reducing storage space requirements, and boosting the performance of later analysis, such as automatic segmentation.

- *Methods*: We propose a SR framework for enhancing images of plant roots by using convolutional neural networks (CNNs). We compare three alternatives for training the SR model: i) training with non-plant-root images, ii) training with plant-root images, and iii) pretraining the model with non-plant-root images and fine-tuning with plant-root images. The architectures of the SR models are based on two state-of-the-art deep learning approaches: i) Fast Super Resolution Convolutional Neural Network; and ii) Super Resolution Generative Adversarial Network.

- *Results*: In our experiments, we observe that the studied SR models improve the quality of the low-resolution images (LR) of plant roots of an unseen dataset in terms of SNR. Likewise, we demonstrate that SR pre-processing boosts the performance of a machine learning system trained to separate plant roots from their background.

- *Discussion*: The incorporation of a deep learning-based SR model in the image formation process boosts the quality of LR images of plant roots. We demonstrate on a collection of publicly available datasets that the SR models outperform the basic bicubic interpolation even when trained with non-root datasets. Also, our segmentation experiments show that high performance on this task can be achieved independently of the SNR. Therefore, we conclude that the quality of the image enhancement depends on the application.

**Key words**: Convolutional neural networks; generative adversarial networks; plant phenotyping; root phenotyping; super-resolution.

## INTRODUCTION

In the last decade, advances in sensing devices and computer systems have allowed for the proliferation of high-throughput plant phenotyping systems (Das Choudhury et al. 2019). These systems are designed to acquire and analyze a large number of plant traits (Han et al. 2014; Krieger 2014), including the measure of small structures, such as the venation network of leaves (Green et al. 2014; Endler 1998). However, the characterization of plant roots is more challenging since they are "hidden" in the soil (Atkinson et al. 2019), which limits the type of sensors and techniques that can be applied.

We categorize the methods that have been used for root analysis as follows: **i) Non-imaging based in-situ methods:** these methods estimate traits of the root system architecture (RSA) by their correlations with chemical or physical properties. For example, in (Cseresnyés et al. 2018; Dalton 1995), the plant root electrical capacitance is used to estimate the root mass (the RSA is modeled as a resistance-capacitance circuit), likewise in (Cao et al. 2011), the electrical impedance spectroscopy (EIS) approach is employed to model the RSA based on the frequency response. The disadvantage of these methods is that they do not provide morphological details since they are a simplified description of the RSA. **ii) Destructive methods:** in this category, we include the techniques that destroy the RSA during or after the imaging process. The most basic of this type is the one called "shovelomics" that consists of washing out the roots of the soil (Trachsel et al. 2011). Shovelomics can be applied in any type of soil, in contrast with other root phenotyping techniques that have limitations regarding the physical properties of the environment. However, it is not ideal for high throughput because the manual excavation of the roots is labor-intensive and tedious. Also, most of the thin roots are lost in this process. **iii) Imaging under controlled conditions:** roots can be observed using rhizotrons, that are structures with windows that contain the soil where the plants are grown (Taylor et al. 1990). Also, 3-D imaging of RSA can be carried out by using special substrates, e.g., transparent substrates or easy-to-remove types of soil (Clark et al. 2011). These procedures allow acquiring high-quality images, but their main disadvantage is that the imaging acquisition is not made in situ. Therefore, the knowledge that can be inferred by them is limited. **iv) Intrusive methods:** this category encompasses the techniques where the acquisition device is introduced into the ground. We consider as intrusive methods the minirhizotrons that use a camera fixed into the soil through a tube to record sequences of pictures of parts of the RSA (Johnson et al. 2001), as well as soil coring (Wu et al.

2018). Although these methods do not necessarily result in the destruction of the RSA, they disturb the roots and soil, which might affect the natural root-soil interactions (Kolb et al. 2017). The disturbance can be worse when the devices are introduced and extracted frequently or when they are installed in stony soils (Majdi 1996). **v) Non-intrusive methods:** these techniques aim to image the RSA in situ, without disturbing the roots or the soil. In (Barton and Montagu 2004), ground-penetrating radar (GPR) technology was tested for this purpose —it was possible to detect tree roots of 1 cm diameter buried in soil at 50 cm depth. However, GPR is currently limited to the detection of roots of trees or woody plants (Araus and Cairns 2014; Hirano et al. 2009). X-ray computerized tomography (CT) (Tabb et al. 2018) and magnetic resonance imaging (MRI) (Pflugfelder et al. 2017) —that consist of scanning by devices traditionally used for medical applications— can be grouped into this category when the complete plant can be scanned  in the device (e.g., plants grown in pots). On the other hand, X-ray CT and MRI based analysis are intrusive when scanning extracted and washed root systems or soil-cores for root architectures removed from the field. In addition to these available approaches, there is on-going development of additional non-intrusive root imaging approaches including backscatter radiography (Cui et al. 2017).

The root system is responsible for water and nutrient absorption, and it is the first barrier to the changing environment. It affects many processes, such as plant growth, $CO_2$ assimilation, and fruit development (Chen et al. 2019; Akinnifesi et al. 1998). Thus, the development of high-throughput root phenotyping methods with low labor inputs is crucial to plant sciences.  As mentioned above, the acquisition of high-resolution (HR) imagery of roots in the field by non-intrusive methods remains a challenge. An effective super-resolution (SR) algorithm that complements the image formation process —by inferring HR details not clearly delineated by the sensing device— is desired for the deployment of these systems in real-world applications.

The SR problem consists of estimating HR images from low-resolution (LR) images. SR has been used to overcome hardware limitations in applications that heavily rely on high-quality images, such as medical diagnosis (Zhang et al. 2012; Zhang and An 2017). Many SR methods in the literature use mathematical transformations of the original data to learn the LR to HR mapping (Yang et al. 2010; Zeyde et al. 2012). For instance, methods based on sparse representations reconstruct each image by a weighted combination of *words* from a set of basic patterns —called a *dictionary*. A set of LR and HR words are learned from training data. A SR image is obtained by replacing the LR dictionary words by HR dictionary words. Recently data-driven SR models based on deep learning algorithms with convolutional neural networks (CNN) have become more popular than the sparse representation-based models. The SR deep learning algorithms are preferred in many cases because they generally exhibit a better performance, additionally they can be applied as a "black box" when enough training data is available (Wang et al. 2015; Ledig et al. 2017). Particularly, super-resolution generative adversarial networks (SRGANs) have shown a high performance on the estimation of HR details loss in a degradation process (Ledig et al. 2017). For the best of our knowledge, SR deep learning models for root imagery have not been extensively studied. Additionally, an effective SR performance measure on this context is unclear since it has been observed in previous studies that reconstruction accuracy (pixel by pixel comparison of a HR-SR pair) and perceptual quality (comparison of visual features of a HR-SR pair) are not directly correlated (Blau et al. 2019).

To enhance plant root imagery, we adapt two state-of-the-art deep learning approaches, the Fast-Super-Resolution Convolutional Neural Network (FSRCNN) proposed in (Dong et al. 2016), and the Super Resolution Generative Adversarial Network (SRGAN). We train the SR models with LR-HR data of two non-root datasets (DIV2K and 91-image) and three plant root datasets (Arabidopsis, Wheat, and Barley). These datasets were selected since they have considerable differences in textures and shapes, which encourage the model to find a general solution. Also, in order to facilitate the training of the generator (the part of the SRGAN that converts LR into HR images), we introduce a modification of the SRGAN by implementing multiple discriminators (the part of the SRGAN that evaluates the quality of the SR images). In the loss function, we consider the mean square error between HR-LR (that reduces the reconstruction error since it is low if the pixel values are similar), and the adversarial loss (that encourages the network to learn to add HR details to the LR image). To evaluate the SR performance, we use two methods: i) computing the standard signal to noise ratio (SNR) between the SR image and the original HR image; and ii) computing the intersection over union (IoU) when applying the SegRoot network (Wang et al. 2019).

The remainder of this paper is divided as follows. In Methods, we describe the models used for training and testing the SR algorithms. In Results, we report the performance of the SR models. In Conclusion, we explain the relevant findings in our study and provide recommendations for the implementation of SR algorithms for root imaging.

## METHODS

In this section, we explain the SR method and the settings that we use to train and test the SR models.

### Super Resolution Model

Many CNN architectures that allow mapping LR images into SR images can be found in the machine learning literature. In this effort, we use two state-of-the-art CNN-based models, FSRCNN and SRGAN. FSRCNN is a model that exhibits similar performance to other state-of-the-art SR techniques, but its execution is considerably faster —this characteristic makes it convenient for comparing different training datasets. Appendix 1 contains a description of the parts of this network. SRGAN is a machine learning system formed by two blocks, a discriminator $D$ and a generator $G$. The function of $D$ is telling apart SR images and real HR images. On the other hand, *G* aims to generate SR images capable of fooling *D*. In Appendix 2, we describe the SRGAN model in detail.

For evaluation purposes, we apply an automatic segmentation on the SR images and quantitatively evaluate the performance of the segmentation. Several U-net encoder-decoder architectures have been proposed for automatic detection and segmentation of plant roots (Xu et al. 2020). In this work, we rely on the SegRoot model. Figure 1 shows the stages of the SR framework applied to enhance plant roots.

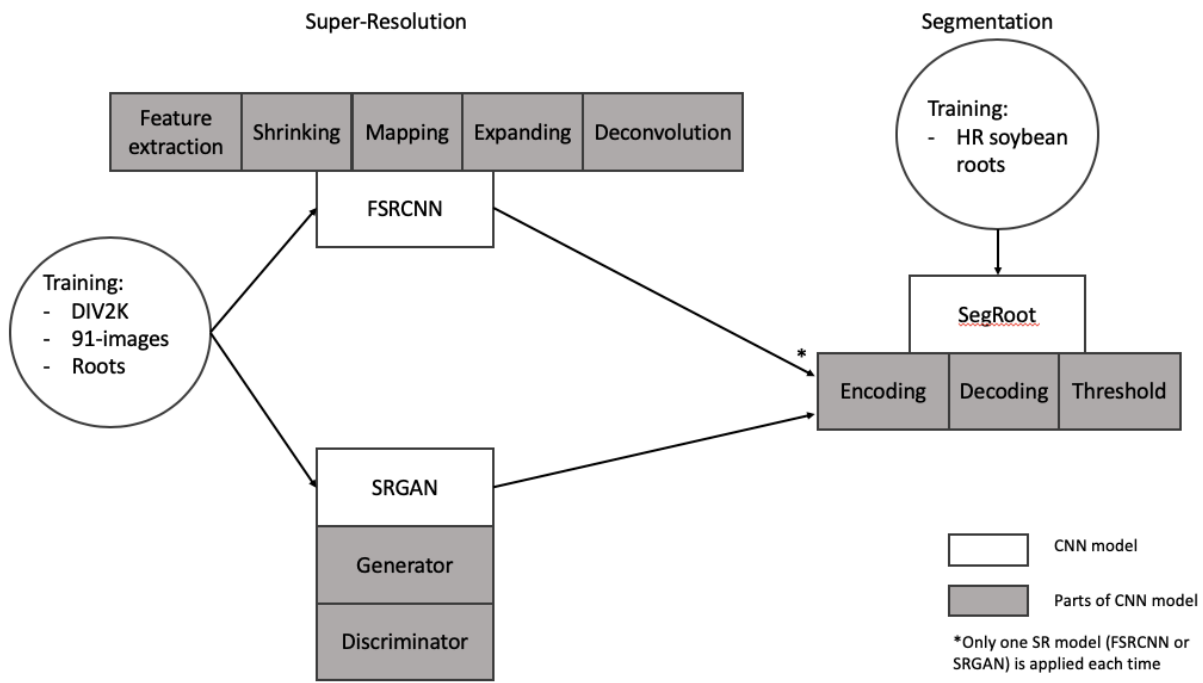

**Figure 1.** Stages of SR experiments. Left: SR models and their parts, FSRCNN and SRGAN. Right: Segmentation model, SegRoot.

## Evaluation

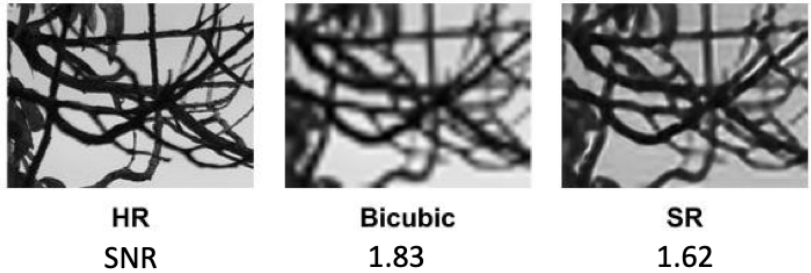

|         |         |         |
|---------|---------|---------|
| **HR**  | **Bicubic** | **SR** |
| SNR     | 1.83    | 1.62    |

**Figure 2.** The SNR and the perceptual quality of an image are not always directly correlated.

We quantitatively evaluate the SR performance by two measures: SNR and IoU. SNR is a classic measure for estimating the quality of a recovered signal. It is computed by a pixel-by-pixel comparison between the original HR image and the estimated SR image, as follows

$$SNR = 10 \, log \, \frac{1}{\|HR-SR\|^2} \, .$$

However, SNR might not necessarily highlight any HR detail enhancement. For example, in Fig. 2, the SNR (the higher the better) of the image estimated by bicubic interpolation (1.83) is higher than the SNR of the SR image (1.62) —even though the interpolated image looks blurred. For this reason, we also estimate the effect of applying the SR enhancement as a preprocessing step in an automatic root-to-background segmentation process. To this end, we trained the state-of-the-art SegRoot network (Wang et al. 2019) with HR data. Therefore, we assume that the segmentation would be more accurate if the input data contains HR details as the ones used for training. We compare the binary ('1'-pixels indicate root, and '0'-pixels indicate background) segmented images $B_{seg}$ with manually labeled images $B_{gt}$ by the IoU (Rahman and Wang 2016), also known as the Jaccard Index computed by

$$IoU = 2 \, \frac{\left| B_{seg} * B_{gt} \right|}{\left| B_{seg} \right| + \left| B_{gt} \right|}$$

where '$|\cdot|$' denotes the sum of all the entries of the input matrix and '$*$' is a pointwise multiplication. The IoU values are between 0 and 1 (the higher the better). An IoU value of '1' is when all the target pixels are correctly classified and there are not false positives.

**Datasets**

In this study, we use five publicly available datasets (two non-plant root datasets and three plant-root datasets) to train the SR models. They are listed below:

- DIV2K (https://data.vision.ee.ethz.ch/cvl/DIV2K/): a dataset of natural images that has been used by others to train and test SR algorithms. We train our models on the grayscale version of the training dataset (800 images).

- 91-Image (https://www.kaggle.com/ll01dm/t91-image-dataset): a classical dataset commonly used in SR studies.

- Arabidopsis roots (https://zenodo.org/record/50831#.XjIAPVNKhQI): an Arabidopsis dataset for root phenotyping analysis (Bouché et al. 2016).

- Wheat roots (http://gigadb.org/dataset/100346): a dataset consisting of 2614 images of wheat seedlings (Atkinson et al. 2017).

- Barley roots: a set of 3-D magnetic resonance images (MRI) of barley roots (https://www.plant-image-analysis.org/dataset/3d-magnetic-resonance-images-of-barley-roots). This dataset also contains WinRHIZO images of the barley roots.

In our experiments, we group the three plant-root datasets (Arabidopsis, Wheat, and Barley) into one that we call "Roots". Figure 3 shows examples of the plant-root datasets used for training the SR model. To test the performance of the SR models, we use the Soybean roots (https://github.com/wtwtwt0330/SegRoot) dataset that consists of 65 images of soybean roots (Wang et al. 2019).

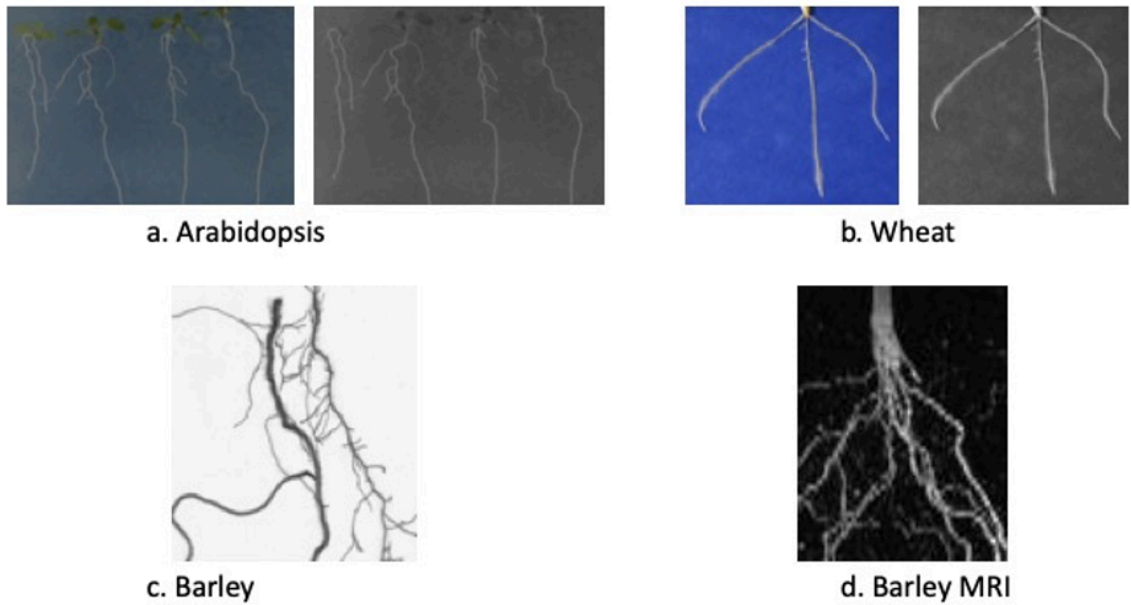

**Figure 3.** Examples of plant-root images used to train SR models. RGB images of Arabidopsis roots (a) and wheat roots (b) were converted to grayscale.

## RESULTS

In our experiments, we train nine SR models: 1) FSRCNN-DIV2K: FSRCNN trained with the DIV2K dataset. 2) FSRCNN-91-image: FSRCNN trained with the 91-image dataset. 3) FSRCNN-roots: FSRCNN trained with the Roots datasets. 4) FSRCNN-91-image&roots: FSRCNN-91-image model fine-tuned with the Roots dataset. 5) SRGAN-DIV2K: SRGAN trained with the DIV2K dataset. 6) SRGAN-91-image: SRGAN trained with the 91-image dataset. 7) SRGAN-roots: SRGAN trained with the Roots dataset. 8) SRGAN-91-image&roots: SRGAN-91-image model fine-tuned with the Roots dataset. 9) SRGAN-MULDIS: SRGAN model trained with three discriminators (one for each dataset: DIV2K, 91-image, and Roots).

For all the SR training experiments, we use as validation dataset a subset of the 100 images from the Roots dataset. The validation dataset is used to estimate the performance of the model in terms of the SNR after completing each iteration. After finishing the training process, we take as the parameters of the model the ones that output the highest SNR on the validation test. Each model is trained 100 iterations (the loss function converges with this number of iterations).

To evaluate the performance of the SR models, we downscale the images of the Soybean dataset by four. We use one of the SR models listed above to upscale the test images to their original resolution. We estimate the SNR by comparing the estimated SR images and original HR images. Afterward, we use the SegRoot network to classify each pixel in the input image as root or non-root. As lower and upper bounds, we take the upscaled images by bicubic interpolation, and the original HR images, respectively. Table 1 contains the SNR and IoU obtained on the grayscale-Soybean dataset. Segmentation carried out on HR always exhibits the best performance. We infer that the HR details on the images boost the performance of the SegRoot model on this data. Also, all the SR models outperform the bicubic interpolation in terms of both SNR and IoU. Regarding only the SR models, three of them, FSRCNN-91-image, FSRCNN-roots, and SRGAN-MULDIS exhibit the highest SNR (we consider that there is not statistical evidence to prefer one of them in this case since their standard error overlaps). However, there is a mismatch between SNR and IoU results. The model that performs the best in terms of the IoU is FSRCNN-91-image&roots. Therefore, the features enhanced by the SR models that allows increasing the SNR do not necessarily imply that can be useful for any task, as the applied automatic segmentation. Figure 4 contains examples of SR and segmented.

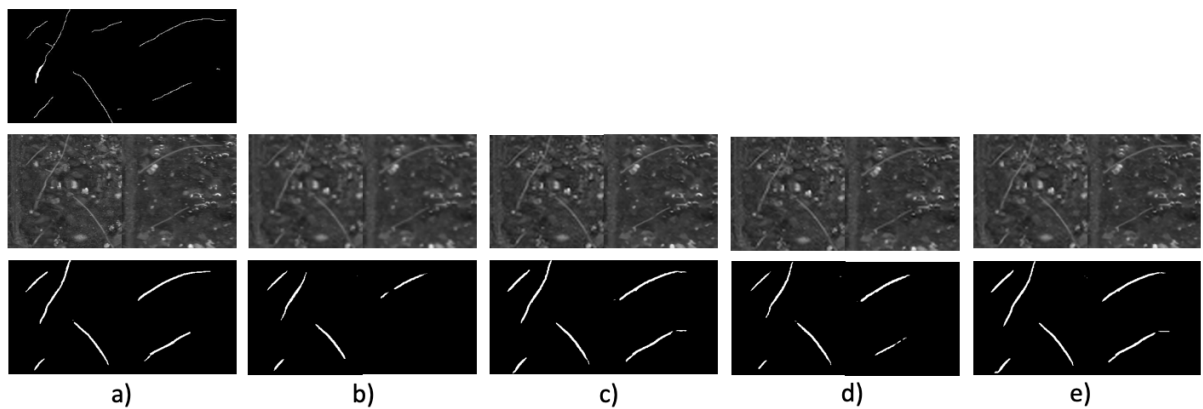

a)          b)          c)          d)          e)

**Figure 4.** SR and segmentation example images (128x64) on the soybean dataset. From top to bottom: a) ground-truth, HR image and segmentation on HR image; b) Bicubic image and its segmentation; c) FSRCNN-91-image model and its segmentation; d) SRGAN-MULDIS model and its segmentation; and e) FSRCNN-91-image&roots model and its segmentation.

Table 1. Evaluation of SR models on Soybean dataset. SNR and IoU mean (and standard error in parentheses)

| Model | SNR | IoU |
|---|---|---|
| 1. Bicubic | 28.30 (1.37) | 0.0984 (0.0098) |
| 2. FSRCNN-DIV2K | 32.60 (0.19) | 0.1313 (0.0106) |
| 3. FSRCNN-91-image | **33.10 (0.20)** | 0.1419 (0.0108) |
| 4. FSRCNN-roots | **33.05 (0.20)** | 0.1623 (0.0111) |
| 5. FSRCNN-91-image&roots | 32.48 (0.19) | **0.1709 (0.0110)** |
| 6. SRGAN-DIV2K | 32.48 (0.19) | 0.1402 (0.0106) |
| 7. SRGAN-91-image | 32.47 (0.19) | 0.1327 (0.0107) |
| 8. SRGAN-roots | 32.71 (0.19) | 0.1485 (0.0108) |
| 9. SRGAN-91-image&roots | 32.66 (0.20) | 0.1536 (0.0108) |
| 10. SRGAN-MULDIS | **33.05 (0.20)** | 0.1415 (0.0108) |
| 11. HR | --- | 0.2003 (0.0122) |

The average processing time of a 64x64 image is 0.2248 s with SRGAN-based models, 0.2170 s with FSRCNN models, and 0.0003 s with bicubic interpolation. Note that the bicubic interpolation is an upscaling method that does not require training. All the computational experiments were performed in a Linux Centos 7 machine, x86_64, Intel Xeon CPU @3.60 GHz with a GPU GeForce RTX.  For implementation, we used the deep learning framework PyTorch 1.2.0.

**DISCUSSION**

We design a framework for the application of deep learning-based SR models to enhance plant root images. In our experiments, we evaluate the SR models in terms of both the reconstruction capability (by SNR), and the boosting of the images for performing automatic segmentation (by IoU). We demonstrate that the SR models outperform the basic bicubic interpolation even when trained with non-root datasets. Also, our segmentation experiments show that a high performance on this task can be achieved independently of the SNR. Therefore, we conclude that the quality of the image enhancement depends on the application.

The pipeline of image processing might include other stages, such as denoising and contrast enhancement. To incorporate any new stage, we recommend using the two-section evaluation method that we applied on this paper: evaluation of the performance directly on the processed image, and evaluation of the result when performing a machine learning task on the processed image. Also, we suggest using SR models for improving the performance of other machine learning tasks such as feature extraction and classification.

Future work could include the application of the proposed SR framework on images acquired on the field. In this work, we generate LR samples by downscaling the original HR images. Therefore, an extension of this work might consider using an alternative to transform HR into LR images, such as "blind SR kernel estimation" methods.

## ACKNOWLEDGMENTS

This work was funded by the Advanced Research Projects Agency-Energy (ARPA-E) under Award Number P0056459. The views and opinions of authors expressed herein do not necessarily state or reflect those of the United States Government or any agency thereof.

## Appendix

**Appendix 1.** Fast Super-Resolution Convolutional Neural Network (FSRCNN)

FSRCNN is divided into five parts: 1) Feature extraction: FSRCNN consists of a convolutional layer with *d* filters of size 5x5 and 1 input channel. In this case, *d* is considered as the LR dimension. This part is denoted by *Conv(5,d,1)*. 2) Shrinking: The purpose of a shrinking layer is reducing the LR dimension. The shrinking procedure is carried out by a convolutional layer of *s* 1x1-filters, denoted by *Conv(1,s,d)* where *s* is smaller than the number of input channels *d*. 3) Mapping: The mapping layer is a non-linear mapping that aims at estimating a shrunken version of the HR dimension. This layer is implemented as a sequence of *m* 3x3 convolutional layers. The number of filters is *s* for each layer. This mapping is denoted by *mxConv(3,s,s)*. 4) Expanding: The expanding layer is implemented using a number *d* of SR feature maps are estimated by a 1x1 convolutional layer (denoted by *Conv(1,d,s))*. 5) Deconvolution: The deconvolution component corresponds to a 9x9 deconvolution layer with 1 filter that upscales *n* times the height-and-width input dimensions. The deconvolution components denoted by *DeConv(9,1,d)*.

As suggested by the authors of FSRCNN, we apply a Parametric Rectified Linear Unit (PReLU) after each convolutional layer. Also, we set the parameters *d*, *s*, and *m* as 56, 12, and 4, respectively. It has been experimentally demonstrated that these values are suitable for recovering HR details.

**Appendix 2.** Super-Resolution Generative Adversarial Network (SRGAN)

A generative adversarial network (GAN) is formed by two blocks, a generator *G* and a discriminator *D*. In this configuration *G* and *D* play contrary roles, i.e., on the one hand, *G* aims at generating ``realistic-like fake data'' capable of fooling *D*, whereas *D* is continuously trained to classify fake data from real data (Goodfellow et al. 2014). Mathematically, the adversarial setting is formulated as follows

$$\min_{G} \max_{D} \; \mathbb{E}_{x \sim p(x)} \log(D(x)) + \mathbb{E}_{x \sim q(x)} \log(1 - D(G(x))) \tag{1}$$

where *x* denotes a sample (e.g., an image), and *p* and *q* are data distributions (e.g., distributions of LR and HR images). Since this is a min-max problem, the expression in (1) is both, a loss function and a reward function. The optimization problem is solved in an alternating manner. In one step, the loss function is minimized w.r.t. *G*, such that the output of *G(x)|x~p(x)* is optimized when *D(G(x))* equals to 1. On the other hand, the expression in (1) is seen as a reward function that is maximized w.r.t. *D*. In this case, *D(x)* is a classifier that is trained to output one when *x~p(x)*, and zero when *x~q(x)*.

To choose the architecture of *D* and *G*, we need a two-class classification network, and a network that outputs a matrix of the same size of the input (since the LR image is interpolated to the size of the desired SR image), respectively. We evaluated several architectures and selected two for their balance between performance and computational requirements. As a *G* network, we use the convolutional super-resolution layers of the resolution-aware convolutional neural network (RACNN) proposed in (Cai et al. 2019). For *D*, we design a two-class classifier with three convolutional layers and one fully connected layer. For training, we use a batch size of 100, and as an update rule, we apply Adaptive Moment Estimation (Adam), applied also in the method proposed by (Ledig et al. 2017), with a learning rate of 0.001. To create LR training images, we randomly select *64x64* chunks, downsample them to *16x16* size and upsample them again to the original size by bicubic interpolation.

In a SRGAN, *x~p(x)* is a sample of a set of LR images, and *x~q(x)* is a sample of a set of HR images. After several iterations, it is expected that the *D* is not able to tell apart HR and SR images, i.e., *G* learns to convert LR images into SR images very similar to the original HR images. Note that in (1), it is not required that the output of the generator matches the HR version of the LR input, i.e., the content of the generated image might not be the same as the one in the LR one. To enforce the matching between HR-LR pairs, we add the squared error between the HR and SR images to the function as follows

$$\min_{G} \max_{D} \quad \mathbb{E}_{x,y \sim p(x,y)} \quad \log(D(x)) + \log(1 - D(G(y))) + \|x - G(y)\|^2 \qquad (2)$$

where $x,y \sim p(x,y)$ is a pair of HR (x) and LR (y) images randomly sampled from a set of HR-LR image pairs, and $\|\cdot\|$ denotes the Euclidean norm. Note that in (2), when minimizing w.r.t. *G,* two terms are considered, $\log(1 - D(G(y)))$ and $\|x - G(y)\|^2$, which correspond to the generative adversarial loss and content loss, respectively (Ledig et al. 2017).

We modify the approach described in (2) by incorporating multiple $D_i$ discriminators in the SRGAN architecture (one discriminator per dataset). Therefore, each discriminator acts as an expert to distinguish HR-SR images on one type of data. The optimization problem is written as follows

$$\min_{G} \max_{D} \sum_{i} \mathbb{E}_{x,y \sim p_i(x,y)} \log\Big(D_i(x)\Big) + \log\Big(1 - D_i(G(y))\Big) + \|x - G(y) \quad (3)$$

where $D_i$ is the discriminator specialized in the *i*-th dataset. We hypothesize that in this way the generator will output more general SR images since it is more challenging to "cheat" several specialized discriminators (one per type of data) than a general one (a discriminator that distinguishes HR vs. SR images of any kind).